\documentclass[10pt]{article} 
\usepackage{cogsci,epsfig,authordate1-4,ipa}

\title{Where Defaults Don't Help: the Case of the German Plural
System}
\author{
{\Large \bf Ramin~Charles~Nakisa \and Ulrike~Hahn} \\
Department of Experimental Psychology, Oxford University \\
South Parks Road \\
Oxford, OX1 3UD \\
{\tt \{ramin,ulrike\}@psy.ox.ac.uk}
}

\begin{document}

\maketitle

\begin{abstract}

The German plural system has become a focal point for conflicting
theories of language, both linguistic and cognitive.  We present
simulation results with three simple classifiers -- an ordinary
nearest neighbour algorithm, Nosofsky's `Generalized Context Model'
(GCM) and a standard, three-layer backprop network -- predicting the
plural class from a phonological representation of the singular in
German. Though these are absolutely `minimal' models, in terms of
architecture and input information, they nevertheless do remarkably
well. The nearest neighbour predicts the correct plural class with an
accuracy of 72\% for a set of 24,640 nouns from the CELEX database. With
a subset of 8,598 (non-compound) nouns, the nearest neighbour, the GCM
and the network score 71.0\%, 75.0\% and 83.5\%,
respectively, on novel items. Furthermore, they outperform a hybrid,
`pattern-associator + default rule', model, as proposed by Marcus et
al. \shortcite{MARCUS95}, on this data set.

\end{abstract}

\section{Introduction}

The German plural system has been the subject of a wide variety of
theoretical accounts ranging from traditional `Item and Process
accounts' \cite{MUGDAN77} to schema theories (Bybee
\shortcite{BYBEE95}; K\"opcke \shortcite{KOEPCKE88,KOEPCKE93}) and
recent `default rule + pattern associator' accounts
\cite{MARCUS95}. Furthermore, it has been championed as a crucial test
case \cite{MARCUS95} in the debate on the psychological reality of
linguistic rules triggered by Rumelhart and McClelland's
\shortcite{RUMELHART86} model of the English past tense.

Decision between the various theoretical accounts is, at present,
difficult; though they have the virtue of dealing with a wide range of
phenomena, they are not explicit enough to allow suitably fine-grained
evaluation. Extant computational models, on the other hand, neither
deal with the German plural\footnote{with the exception of Goebel and
Indefrey \shortcite{GOEBEL94}.} nor attempt to capture the full range
of phenomena such as pluralization of truncations, acronyms, quotes
etc. compiled by Marcus et al. \shortcite{MARCUS95}. What is now
required is the development of explicit computational models which
allow quantitative assessment against real data. As a starting point,
we have implemented and tested three `minimal models' -- simple,
off-the-shelf classifiers -- which, given phonological information
about the singular alone, predict the correct plural class with
surprising accuracy. These are not advanced as full-blown cognitive
models of the German plural, but rather as benchmarks against which
more complex accounts must be compared. As an example of this, we also
pitted these models against three versions of a hybrid, `associative
memory+default rule' model \cite{MARCUS95}, which subsumes them in the
associative component.

\section{The Task}

{\bf The Data Sets} Our dataset is drawn from the 30,100 German nouns
in the CELEX database.\footnote{CELEX can be obtained by contacting
celex@mpi.nl.} Since the CELEX classification is fraught with error,
we automatically classified nouns according to the nature of the
transformation from singular to plural phonology. Four general types
of transformation occur: identity mappings, suffixation, umlaut (vowel
change) and rewriting of the final phoneme(s). The classification
yields approximately 60 categories (some of which contain only one
member).

We then discard categories with a type frequency of less than 0.1\%
resulting in a database of 24,640 nouns with 15 different plural
categories (see table \ref{typefreqs}).\footnote{45 words and 2
duplicates were manually removed because they were obviously incorrect
(e.g. incorrectly pluralized proper names and entries with errors in
phonological form).} This step removes primarily latinate and Greek
words and a small number of German words with arbitrary plurals
(suppletion, or singly occurring transformations). In effect this
brings our classification into accord with the plural types described
in standard linguistic analysis \cite{KOEPCKE88}. The only further
amendment in this direction was that the umlauts (\"a, \"o, \"u) were
treated as one as is consensual in the literature.

For computational reasons this set was further reduced to a set of
8,598 ``non-compound'' nouns. A ``non-compound'' noun was defined as a
noun that did not contain another noun from the database as its
rightmost lexeme.\footnote{This leaves complex nouns which are not
noun-compounds, and noun-compounds for which the right-most lexeme is
not listed individually.} This is justified by the fact that, in
German, the plural of a noun-compound is determined exclusively by the
right-most lexeme, making the remainder of the word redundant. That
this reduction does not distort the similarity structure of the German
lexicon is borne out by the fact that the performance of the nearest
neighbour classifier on the entire data set and the subset were
virtually identical (72\% and 71\% respectively). This dataset of
8,598 nouns was split roughly in half to give a training set (4,273
words) and a testing set (4,325 words). A copy of the training set
which had all 282 words that took a +s suffix removed (leaving 3,991
training words) was used to train the hybrid rule-associative models.

\begin{table*}[htbp]
\begin{center}
\begin{tabular}{||l|r|r@{.}l|r|r@{.}l||} \hline \hline
\bf Plural & \multicolumn{3}{c|}{\bf All Nouns} & \multicolumn{3}{c||}{\bf Non-Compound Nouns} \\
\bf Type & \multicolumn{1}{l}{\it Frequency} & \multicolumn{2}{c|}{\it \% of Total} &\multicolumn{1}{l}{\it Frequency} & \multicolumn{2}{c||}{\it \% of Total} \\
\hline \hline
+{\schwa}n                          &  7012 & 28&109 & 2646 & 30&775 \\
+n                                  &  4477 & 17&947 & 1555 & 18&086 \\
+\schwa                             &  4460 & 17&879 & 1178 & 13&701 \\
Identity                            &  4201 & 16&840 & 1992 & 23&168 \\
Umlaut+\schwa                       &  2017 &  8&085 &  239 &  2&780 \\
+s                                  &   978 &  3&920 &  571 &  6&641 \\
Umlaut+{\schwa}r                    &   692 &  2&774 &   54 &  0&628 \\
+{\schwa}r                          &   289 &  1&159 &   36 &  0&419 \\
Umlaut                              &   255 &  1&022 &   35 &  0&407 \\
{\niupsilon}m$\rightarrow${\schwa}n &   135 &  0&541 &   95 &  1&105 \\
a$\rightarrow${\schwa}n             &   121 &  0&485 &   81 &  0&942 \\
{\niupsilon}s$\rightarrow${\schwa}n &    88 &  0&353 &   69 &  0&803 \\
{\niupsilon}m$\rightarrow$a         &    45 &  0&180 &   40 &  0&465 \\
+t{\schwa}n                         &    27 &  0&108 &    1 &  0&012 \\
+i{\schwa}n                         &    25 &  0&100 &    6 &  0&070 \\ \hline\hline
\end{tabular}
\caption{\label{typefreqs} Frequencies of different plural types in
the complete set of nouns in CELEX and for the non-compound
nouns. Suffixation is indicated by {\em +suffix}, rewrites are
indicated as {\em``phonemes''$\rightarrow$``phonemes''}.}
\end{center}
\vspace*{-0.1in}
\end{table*}

{\bf Input Representation} A phonological representation of the nouns
was created by taking the phonetic singular and plural forms of each
word as given in the CELEX database and rewriting them as a bundle of
15 phonetic features taken from Wurzel \shortcite{WURZEL81}. Sixteen
phonetic slots were used, so each word was represented as a vector
with 240 elements. Since words vary in length, their representations
must be zero-padded. Vectors were right-justified since word endings
are most salient for determining the plural type of German
nouns.\footnote{This was determined by comparison of performance on
left-justified, centre-justified and right-justified words using ID3
\cite{QUINLAN92}.}

{\bf Output} In all cases a model was required to produce the correct
plural category for a given input. Two of the models (GCM and the
network) produce graded responses (probabilities or activations). The
single highest probability/activation was taken to be the output. Only
exact matches were scored as correct. In the following simulations the
simple pattern classifiers were trained on the training data set of
4,273 non-compound nouns. Performance was then assessed on the test
data set of 4,325 words.

\section{Associative Model Performance}

{\bf Nearest Neighbour Classifier} A nearest neighbour classifier
simply adopts the classification of the item in memory most similar to
the new item. It is the simplest kind of exemplar model. In linguistic
terms, it constitutes a `weak analogy' model in K\"opcke's
\shortcite{KOEPCKE88} sense. It could also straightforwardly form the
heart of an associative memory system \cite{PINKER93}; all that would
be required in addition is a component which generates the appropriate
phonological form according to the computed plural class. The nearest
neighbour algorithm simply states: for a novel exemplar $\vec{e}$,
find the most similar (nearest) neighbour $\vec{n}$ and adopt its
plural class. As a similarity metric we used Euclidean
distance. Tested on the 4,325 word testing set, the nearest neighbour
classifier scored 71\%. The pattern of errors, which was basically the
same for all models, reveals an interaction between type-frequency and
class topology. Generally peformance declines with dropping frequency,
but particular low frequency classes can nevertheless be classified
very accurately.\footnote{The network deviates slightly from this
pattern insofar as the lower frequency classes -- from
{\niupsilon}m$\rightarrow${\schwa}n onward -- are highly sensitive to
the initial random seed, so that perfomance can vary drastically between
networks.} In addition, this classifier was also tested on the entire
data set. Here, each noun was individually ``removed'' from the data
set and classified according to the remaining nearest neighbours
giving a classification accuracy of 72\%.

{\bf Nosofsky's Generalized Context Model} Nosofsky's well-known
`Generalized Context Model' \cite{NOSOFSKY90}, which accurately fits
human performance data on a range of classification tasks, is a more
sophisticated exemplar-model, providing a probabilistic
response. Here, the strength of making a category J response ($R_J$)
given presentation of stimulus $i$ ($S_i$) is found by summing the
(weighted) similarity of stimulus $i$ to all presented exemplars of
category $J$ ($C_J$) then multiplying by the response bias for
category $J$. The denominator normalises by summing the strengths over
all categories.

\vspace*{-0.1in}
\begin{equation}
\label{nosofsky}
P( R_J | S_i ) = \frac{ b_J \sum_{j \in C_J} L( j, J ) \eta_{ij}}
{\sum_k b_K \sum_{k \in C_K} L( k, K ) \eta_{ik}}
\end{equation}

In equation \ref{nosofsky} $\eta_{ij}$ ($\eta_{ij}=\eta_{ji}$,
$\eta_{ii}=1$) gives the similarity between exemplars $i$ and $j$,
$b_J$ ($0 \le b_J \le 1$, $\sum b_K = 1$) is the bias associated with
category $J$ and $L(j, J)$ is the relative frequency (likelihood) with
which exemplar $j$ is presented during training in conjunction with
category $J$. The distance $d_{ij}$ is scaled and converted to a
similarity measure using the transformation $\eta_{ij} =
\exp-(d_{ij}/s)^p$ where $p=1$ yields an exponential decay similarity
function and $p=2$ gives a gaussian similarity function.\footnote{Bias
  terms were omitted to limit the number of free parameters.} When the
scaling parameter $s$ was optimised for the gaussian similarity
function ($\eta_{ij} = \exp-(d_{ij}/s)^2$) the performance was 75.0\%
($s=1.46$). When optimised for the exponential ($\eta_{ij} =
\exp(-d_{ij}/s$) accuracy was 74.4\% ($s=0.35$). The gaussian
similarity function was used in the following.

{\bf Neural Network} The neural network most directly resembles the
pattern-associator posited as a module necessary for inflectional
morphology by Pinker \shortcite{PINKER93}, and Marcus et
al. \shortcite{MARCUS95}, with one exception; our network classifies
the input as belonging to one of the 15 plural types (see table
\ref{typefreqs}) instead of directly producing the plural form, in
order to allow comparison with the nearest neighbour and the
GCM. For a full model, a component producing this form on the
basis of class must be assumed.

The network was a three-layer, feed-forward network with 240 input and
15 output units. Different numbers of hidden units -- 10, 20, 30, 40
and 50 -- were tried. Training used back-propagation, duration being
varied from 5 to 50 epochs in steps of 5 epochs and using 3 different
initial random seeds. The best set of weights (defined by
generalisation accuracy on the testing set) was used. It was found
that for all numbers of hidden units the score was at roughly 80\%
after 5 epochs and remained above 80\% up to 50 epochs. The accuracy
of the best network (with 50 hidden units and after 35 training
epochs) was 83.5\%.

\section{Comparing Associative and Rule-Associative Models}

\subsection{Defining Interaction of Associative and Rule Components}

The most recent account of the German plural system by Marcus et
al. \shortcite{MARCUS95} argues that {\em +s} is the `regular' plural
in German; it is produced by a (cognitively real) default rule {\em
`add -s'} which is applied whenever `memory fails'. This lexical
memory is thought to include a phonologically-based, possibly
connectionist, pattern-associator as a subcomponent, hence explaining
the limited productivity of the `irregulars'.

The inflection of the `regulars' on this account, is independent of
the lexicon, resulting from the `rule-route'. This suggests a simple
comparison between pattern associators, which treat the `regulars'
like every other group, and a hybrid rule+pattern associator model, in
which the `regulars' are removed from the pattern-associator and
inflected via the rule-route if `memory fails'. As outlined, all three
models above can form the heart of an associative memory system, and,
thus, can be used for such a comparison. This comparison requires that
Marcus et al.'s notion of memory failure must be made computationally
explicit. We did this through the definition of a threshold {\em t},
as follows:

(1) for {\em nearest neighbour} `memory failure' occurs if the nearest
neighbour in the phonological space is at a distance greater than
$t$. In this case the default inflection +s is used.

\vspace*{-0.1in}
\begin{equation}
\label{nnalgorithm}
\begin{array}{l@{\quad}l}
\mbox{if distance}(\vec{e} - \vec{n}) < t & \mbox{inflect as n} \\ &
\mbox{otherwise use default inflection}
\end{array}
\end{equation}

This means, that for very low values of $t$ the nearest neighbour
memory always fails because there is never a neighbour close enough so
that every singular is classified as a +s. For very large values of
$t$ there is always a nearest neighbour closer than $t$ so the default
rule is never used and the singular is classified using the plural
type of its nearest neighbour. In other words, as $t$ increases, the
algorithm in equation \ref{nnalgorithm} asymptotically reverts to the
nearest neighbour algorithm.

(2) In the {\em GCM} memory `fails' if the largest class probability
$P_J$ was less than a threshold value.

\vspace*{-0.1in}
\begin{equation}
\begin{array}{l@{\quad}l}
\mbox{if MAX}(\vec{P_J}) > t & \mbox{inflect as most probable class}
\\
                             & \mbox{otherwise use default inflection}
\end{array}
\end{equation}

$P_J$ is low and memory failure occurs if the noun is surrounded by
roughly equal numbers of two or more classes of noun or is in a
sparsely populated region of the phonological space.

(3) For the {\em neural network}, finally, memory `fails' if the
greatest output unit activity $MAX(\vec{o_i})$ was less than a
threshold value $t$.

\vspace*{-0.1in}
\begin{equation}
\begin{array}{l@{\quad}l}
\mbox{if MAX}(\vec{o_i}) > t & \mbox{inflect as class of most active
unit} \\
                             & \mbox{otherwise use default inflection}
\end{array}
\end{equation}

For testing, we compute values for $t$ throughout the entire interval
($0.0 < t < 1.0$ for GCM and network and $0.0 < t < \infty$ for
nearest neighbour) in search of an optimal value, and compare the
performance of the hybrid with the simple classifier at each point.

\subsection{Rule-Associative Model Performance}

The hybrid models were assessed on the same test set of 4,325 nouns as
the simple classifiers. In these hybrid models, however, the training
set had the +s nouns removed, since the hybrid model requires that
these are dealt with by the rule alone. Performances are compared with
that of the respective simple classifier trained on the set that
included +s nouns.

\begin{figure}[htbp]
\begin{center}
\input{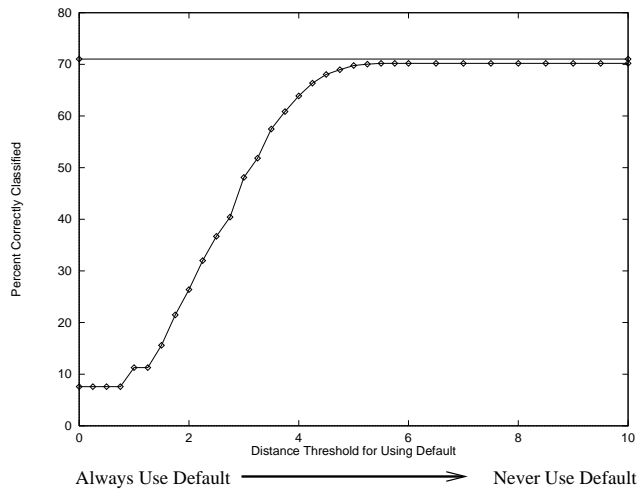}
\vspace*{-0.1in}
\caption{ \label{nndefault} Performance of nearest neighbour
classifier on 4,325 German nouns. Horizontal line shows the
performance of a nearest neighbour classifier with no default rule and
+s plurals included in the training set. Curve shows the performance
of a hybrid rule-associative classifier with +s plurals excluded from
the training set. Increasing use of the default rule from never being
used (threshold = 10) to always being used (threshold = 0)
monotonically reduces classification accuracy.}
\end{center}
\vspace*{-0.1in}
\end{figure}

{\bf Nearest Neighbour Classifier} In order for the addition of a
default rule to improve performance the singular forms of the nouns
that take +s would have to be far away from other singular forms in
sparsely populated areas of the phonological space. However, the
results in figure \ref{nndefault} clearly show that this is not the
case: the classification accuracy increases monotonically with
increasing $t$. In other words, as the frequency of using the default
rule increases from zero, it always deteriorates the performance of
the system. At no value of $t$ does the default improve the
performance above that of the purely associative nearest neighbour
classifier, making the default rule route completely redundant.

{\bf Nosofsky's GCM} The removal of the +s singulars from the training
set changed the optimal value of $s$, so the model was re-optimised
using the training set without +s plurals. The error surface was
sampled in the range $s=1.4$ to $s=1.5$ in steps of 0.01 and $t=0$ to
$t=1.0$ in steps of 0.01. It was found that the optimal value of $s$
was changed slightly to $s=1.48$ and the optimal value of $t$ was
$t=0.29$ giving a classification accuracy of 74.6\%.

\begin{figure}[htbp]
\begin{center}
\input{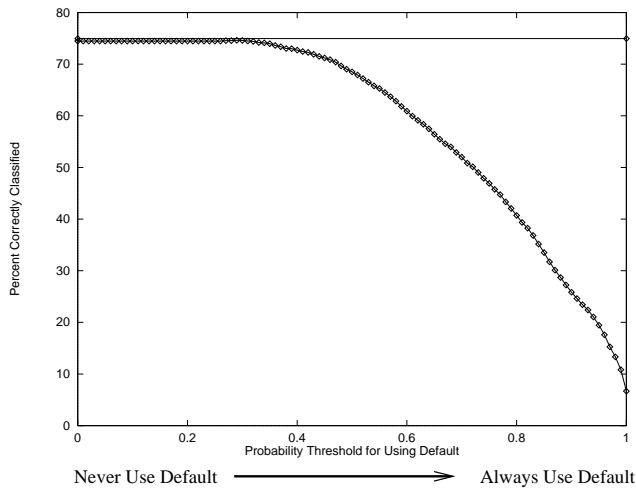}
\vspace*{-0.1in}
\caption{ \label{nosofskydefault} Performance of the GCM on 4,325 German nouns. Horizontal line shows the
  performance of the similarity based classifier with no default rule
  and +s plurals included in the training set. Curve shows the
  performance of a hybrid rule-associative classifier with +s plurals
  excluded from the training set.}
\end{center}
\vspace*{-0.1in}
\end{figure}

Unlike the nearest neighbour, this pattern associator had an optimum
value for the threshold.\footnote{Notice that the threshold value is a
  {\em probability.} for this model, so the performance drops as the
  threshold value increases, whereas for the nearest neighbour the
  threshold was the {\em distance} of the nearest neighbour so that
  performance increased with increasing threshold values.} There was a
0.2\% increase in performance to 74.6\% correct at a probability
threshold of 0.29 from 74.4\% correct at probability threshold 0.0
(see figure \ref{nosofskydefault}). Performance of the
rule-associative classifier never reached that of the purely
associative classifier.

\begin{figure}[htb]
\begin{center}
\input{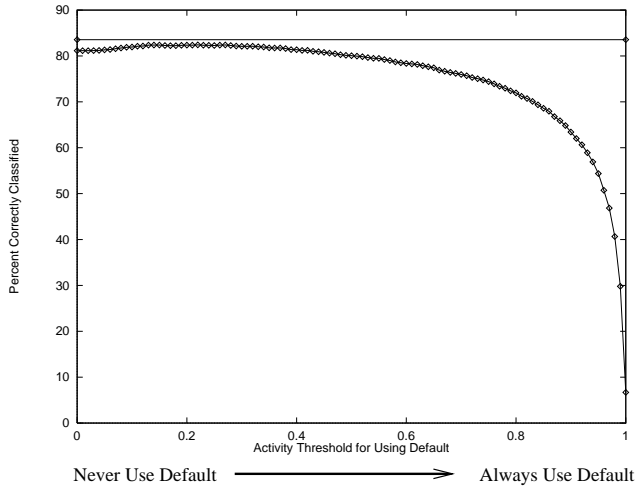}
\vspace*{-0.1in}
\caption{ \label{netdefault} Performance of 3 layer network
classifier on 4,325 German nouns. Horizontal line shows the
performance of the network classifier with no default rule and +s
plurals included in the training set. Curve shows the performance of a
hybrid rule-associative classifier with +s plurals excluded from the
training set.}
\end{center}
\vspace*{-0.1in}
\end{figure}

{\bf Network Classifier} For the rule-associator classifier, the
network was trained on the training set with +s nouns removed and
tested on the standard testing set. Results for this model again
showed a decrease in performance on the addition of a rule (see figure
\ref{netdefault}). There was a 1.2\% increase in accuracy to 82.4\%
correct at an activity threshold of 0.22 from 81.2\% correct at an
activity threshold 0.0. This remained below the 83.5\% accuracy of a
purely associative classifier.

\begin{table}[htb]
\begin{center}
\begin{tabular}[hbtp]{||l|r@{.}l|r@{.}l||} \hline \hline
\it Pattern Associator & \multicolumn{2}{c|}{\it Simple} &
\multicolumn{2}{c||}{\it Hybrid} \\
\hline \hline
Nearest Neighbour                    & 71&0 & 70&2 \\
Nosofsky GCM                         & 75&0 & 74&6 \\
Three-layer Perceptron               & 83&5 & 81&4 \\ \hline \hline
\end{tabular}
\caption{ \label{resultssummary} Summary of associative and
rule-associative model evaluations. The performance of the associative
classifier was greater than that of the hybrid rule-associative
classifiers for all three types of pattern associator.}
\end{center}
\vspace*{-0.1in}
\end{table}

\section{Where a Default {\em Would} Help}

The failure of the hybrid models to outperform the simple models
reflects an important distributional fact about the language.
Performance is never superior because even for the optimum value of
$t$ the rule produces false positives. Increasing performance on the
regulars {\em decreases} the systems performance on the irregulars. This is
because the distances between the regulars are not sufficiently
different from the within-group distances of the irregulars. If they
were, then it would be possible to ``drive a wedge'' between them i.e.
select a value of $t$ that correctly classifies regulars whilst
leaving the irregulars untouched. These considerations suggest that
distributions are possible for which a default {\em would} help.

\begin{figure*}[htb]
\begin{center}
\input{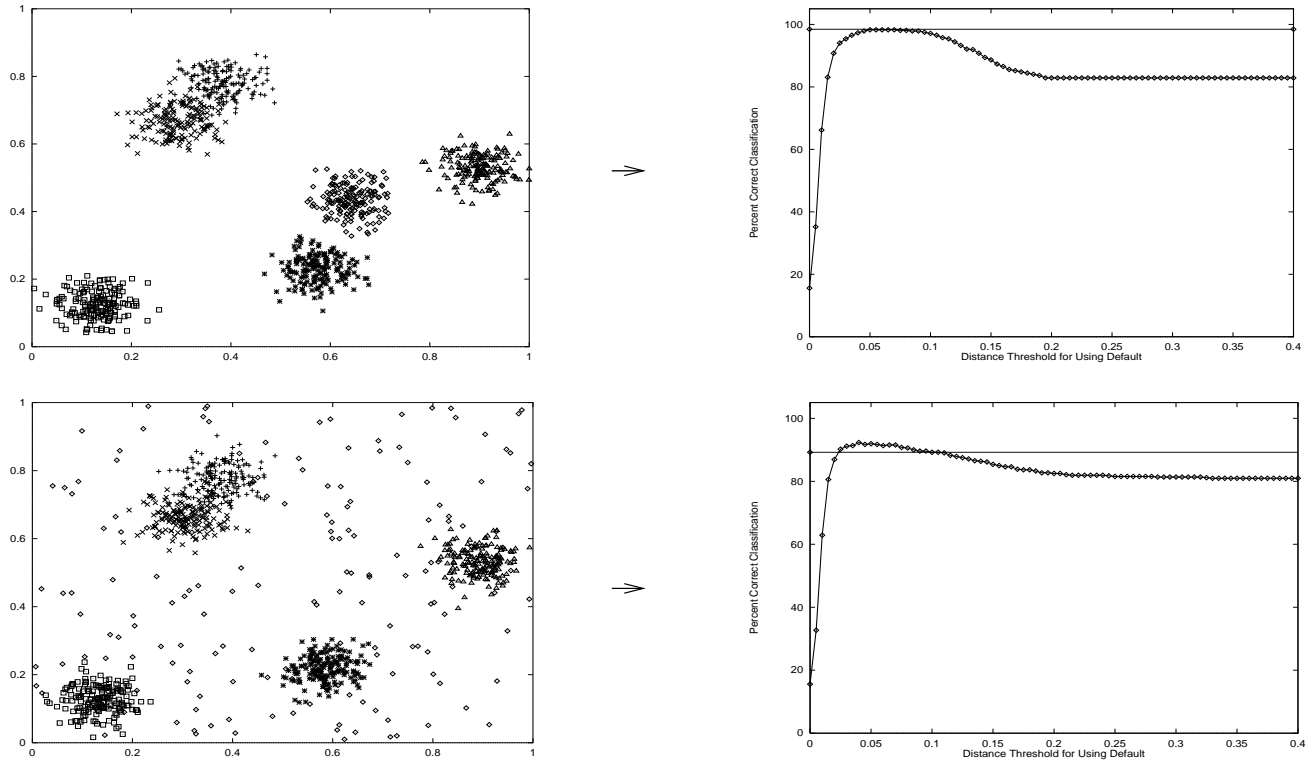}
\caption{ \label{pseudogerman} Two pseudolanguages (left) and their
  corresponding simple and hybrid classifier performances (right). The
  regular class in both languages is shown as diamonds. The top
  language (language 1) has equal variances for all plural types,
  whereas the bottom language (language 2) has the ``regular'' class
  exploded to occupy the entire space homogeneously.}
\end{center}
\end{figure*}

We generated two simple artificial languages to illustrate this. Both
languages consisted of five plural types distributed in a
two-dimensional ``phonological'' space. Each noun class was generated
around a centroid with a gaussian distribution. For the first language,
all five plural types had the same variance, whereas for the second, 
one group, the ``default'', was exploded to occupy the entire space
homogeneously. Both distributions are depicted in figure
\ref{pseudogerman}.

For the first language, where the ``default'' plural type had the same
variance as the other types, the simple nearest neighbour classifier
outperformed the hybrid classifier. By contrast, in the second
language, the hybrid nearest neighbour classifier outperformed the
simple nearest-neighbour classifier. For a distribution where the
irregulars are relatively compact and the regular is homogeneously
distributed, adding a default can be beneficial for generalization.

The default helps by increasing accuracy on a particular subset of the
regulars. It is the regulars forming a shell around each of the
irregular clusters that are correctly classified by the hybrid model
but not by the simple classifier. We call these regulars
``interfacial'' because they are distributed on the surface of the
irregular clusters. Regulars in isolated regions of the space,
``isolated regulars'', are equally well classified by hybrid and
simple models. Thus, increasing the ratio of ``interfacial'' to
``isolated'' regulars increases the benefit of the default. This can
be achieved both by increasing the number of irregular plural types
and (or) by increasing the surface area of irregular plural types.

It is not just the fact that regulars are distributed homogeneously
throughout phonological space \cite{MARCUS95} that
matters, but the existence of interfacial regulars that is
crucial. Isolated regulars alone only allow a threshold $t$ at which
the hybrid's performance is not worse, they do not enable it to do better. In
summary, for particular distributions a default rule can help. Our
results suggest that the German plural system is not of this kind.

\section{Discussion}

It is hard to estimate the maximal score any model performing
prediction could hope to achieve. German has lexical items with
conflicting plural entries and the system as a whole is generally not
presumed to be completely deterministic, allowing a certain degree of
arbitrary exceptions. Whether this means a maximal score should be
placed at 85 or 99\%, the performance of all three purely associative
models seems remarkably high; none are in any way specifically
designed or adjusted for the task, and the input information is
minimal. Other sources of information which have been advanced as
determinants of German plural morphology are semantics \cite{MUGDAN77}
and gender \cite{MUGDAN77,KOEPCKE88}; additionally, syllable
structure, stress and token frequency are likely contributors. Future
work will seek to determine exactly what additional benefits these
sources provide.

For generalisation accuracy on test items drawn from the extant German
lexicon, then, a `default rule' model has no gain whatsoever, and, in
fact, slightly decreases performance (see table \ref{resultssummary}
for summary).\footnote{In terms of the systems confidence, adding a
  default has no visible impact on the
  irregulars. Trivially, confidence on the regulars is increased to
  certainty. Whether this is desirable, is an open psychological question.} Of course, the primary motivation for the `default
rule' account is the fact that it parsimoniously unifies 21 otherwise
seemingly heterogeneous phenomena to which the s-plural is exclusively
or predominantly applied -- such as quotations, acronyms, truncations,
proper names -- \cite{MARCUS95}, which are not captured in our data
set. However, the same threshold {\em t}, which best fits the common
nouns investigated here must also give the right mixtures between
`regular inflection' and `irregulars' for each of the remaining
phenomena. This may or may not be possible; there is no {\em a priori}
reason to believe that it is. This can only be resolved by further
empirical work. In the meantime, these results warn of the way in
which the general, theoretical accounts mentioned in the introduction
are prone to taking computationally consequential details for granted.

\section{Acknowledgements}

The order of the authors is arbitrary. Thanks go to Paul Cairns with
whom this project began, and to Nick Chater who first suggested
nearest neighbour and GCM. Thanks also to Todd Bailey, Kim Plunkett
and two anonymous reviewers for helpful comments on the
manuscript. Ulrike Hahn is supported by ESRC grant
No. R004293341442. Ramin Nakisa is supported by a Training Fellowship
from the Medical Research Council.


\bibliography{german}
\bibliographystyle{authordate1}

\end{document}